\documentclass{article} 
\usepackage{iclr2026_conference,times}

\usepackage{amsmath,amsfonts,bm}

\def\eqref#1{equation~\ref{#1}}

\def\1{\bm{1}}

\DeclareMathAlphabet{\mathsfit}{\encodingdefault}{\sfdefault}{m}{sl}
\SetMathAlphabet{\mathsfit}{bold}{\encodingdefault}{\sfdefault}{bx}{n}

\usepackage{hyperref}
\usepackage{url}
\usepackage{times}
\usepackage{latexsym}
\usepackage{graphicx}
\graphicspath{ {./Figures/} }
\usepackage{amssymb}
\usepackage{multirow}
\usepackage{rotating}
\usepackage[export]{adjustbox}
\usepackage{tikz}
\usepackage{float}
\usepackage{xcolor,colortbl}
\usepackage[most]{tcolorbox}
\usepackage{tabularx}
\usepackage{subcaption}
\usepackage{enumitem}
\usepackage{caption}
\usepackage{booktabs}

\title{An LLM-Based System for Argument Mining}

\author{Paulo Pirozelli\thanks{Corresponding author: \href{ppirozelli@usp.br}{ppirozelli@usp.br}.}, Victor Hugo Nascimento Rocha \& Fabio G. Cozman \\
Universidade de São Paulo\\
Center for Artificial Intelligence (C4AI)
\And
Douglas Aldred \\
Instituto Mauá de Tecnologia\\
Núcleo de Sistemas Eletrônicos Embarcados (NSEE)
}

\iclrfinalcopy 
\begin{document}

\maketitle

\begin{abstract}
Arguments are a fundamental aspect of human reasoning, in which claims are supported, challenged, and weighed against one another. We present an end-to-end large language model (LLM)-based system for reconstructing arguments from natural language text into abstract argument graphs. The system follows a multi-stage pipeline that progressively identifies argumentative components, selects relevant elements, and uncovers their logical relations. These elements are represented as directed acyclic graphs consisting of two component types (premises and conclusions) and three relation types (support, attack, and undercut). We conduct two complementary experiments to evaluate the system. First, we perform a manual evaluation on arguments drawn from an argumentation theory textbook to assess the system's ability to recover argumentative structure. Second, we conduct a quantitative evaluation on benchmark datasets, allowing comparison with prior work by mapping our outputs to established annotation schemes. Results show that the system can adequately recover argumentative structures and, when adapted to different annotation schemes, achieve reasonable performance across benchmark datasets. These findings highlight the potential of LLM-based pipelines for scalable argument mining.
\end{abstract}

\section{Introduction}
Arguments are developed to justify decisions, reason through problems, and persuade others of particular stances \citep{eemeren2018argumentation}. In making an \textit{argument}, a person offers reasons to support or challenge a claim with the aim of providing a rational basis for accepting or rejecting it. By argument, we mean the product of this process: a set of propositions in which one proposition (the conclusion) is claimed to be supported or challenged by others (the premises). These premises provide reasons intended to justify or refute a conclusion that is not immediately evident \citep{hoffmann2023bad}.

Given their central role in reasoning and decision-making, it is not surprising that artificial intelligence (AI) has long sought to model and assess arguments computationally. Traditionally, this goal has relied on \textit{argumentation frameworks} (AFs) \citep{dung1995acceptability}, which provide formal representations for determining whether an argument stands—that is, whether it is supported by sufficient reasons and not successfully undermined by counter-reasons (or whether such counter-reasons are themselves defeated). However, applying AFs to natural language remains difficult: key premises are often implicit in context, either because they are taken as obvious or because speakers prefer not to expose them to scrutiny. Moreover, while AFs are well suited for evaluating the overall acceptability of arguments in a debate, they are typically less equipped to assess the plausibility of individual premises and the strength of specific inferential links, which often depend on commonsense knowledge and pragmatic factors that escape purely formal treatments.

The recent growth of large language models (LLMs), such as OpenAI’s GPT models \citep{singh2025openaigpt5card} and Meta's Llama models \citep{touvron2023llama}, represents a promising development for computational argumentation. These models capture lexical, syntactic, and pragmatic regularities, and can uncover implicit connections that are difficult to formalize. Nonetheless, while LLMs can be effective in downstream tasks such as argumentative span detection and relation classification, they do not by themselves provide a systematic structure for representing and evaluating arguments. In particular, these tasks remain underspecified in terms of what should count as an argument and which criteria should guide its evaluation. Thus, off-the-shelf generations from LLMs are not a substitute for structured representations: without an explicit schema and controllable intermediate steps, outputs can be inconsistent and difficult to compare or aggregate across documents and datasets.

Hence, both AFs and LLMs offer valuable strengths as well as shortcoming. AFs provide explicit and well-defined representational structures, while LLMs offer rich contextual and implicit knowledge derived from large-scale language data; at the same time, AFs methods often struggle with the complexity and variability of real-life arguments, whereas LLM-based approaches may lack systematic and theoretically grounded standards for argument mining. 

To bridge these perspectives, we combine these approaches into a unified framework for reconstructing arguments from natural language text. We employ a multi-stage LLM-based pipeline that identifies argumentative components and infers their relationships, representing arguments as graphs. To ensure well-formed outputs, we incorporate preprocessing steps that enforce a consistent graph structure.

Our main contributions are:
\begin{itemize}
    \item An end-to-end LLM-based system for reconstructing arguments as graph structures from natural language text;
    \item A small annotated argumentation dataset derived from an argumentation theory textbook.\footnote{Code and data are available at \url{https://github.com/do-ald533/llm_argumentation}.}
\end{itemize}

\section{Related Work}
Research on argumentation has developed along two main directions: (i) formal models of argument representation and (ii) methods for extracting arguments from natural language text. Formal models, particularly argumentation frameworks, provide abstract representations and semantics for reasoning about argumentative relations such as conflict and support. Argument mining focuses on identifying argumentative components and relations in raw text. These strands are often pursued independently, leaving a gap between formal representations and natural language approaches to argument extraction.

\subsection{Argumentation Framework}
Argumentation frameworks provide an abstract, graph-based model in which arguments are represented as nodes and conflicts between them as a binary attack relation. Semantics then determine which sets of arguments can be jointly accepted (e.g., grounded, preferred, and stable extensions) \citep{dung1995acceptability,baroni2011semantics}. Several extensions increase expressivity, including value-based AFs, bipolar AFs, and abstract dialectical frameworks \citep{benchcapon2003value,cayrol2005bipolar,brewka2010adf}. Structured formalisms such as ASPIC+ and assumption-based argumentation reconnect abstract arguments to premises and inference rules, enabling explanations in terms of different kinds of defeats \citep{modgil2014aspic,dung2009assumption}. 

Despite their success, AF approaches typically assume that arguments and relations are already given, leaving their extraction from raw text outside the scope of the framework. Moreover, evaluating the plausibility of individual premises or inferential steps often requires commonsense and pragmatic knowledge that formal semantics alone cannot capture. Our system addresses these limitations by using LLM-based modules to recover arguments directly from natural language while imposing AF constraints.

\subsection{Argument Mining}
Argument mining seeks to recover argumentative structure from natural language—typically by segmenting text into argumentative discourse units, identifying their roles (e.g., major claim, claim, premise), and predicting the relations of support or attack that connect them. Early work in AM relied on feature-rich statistical models with handcrafted features. Later neural approaches replaced these with pre-trained encoders, leading to substantial gains in generalization and robustness \citep{lippi2016argumentation, lawrence2020argument}. 

More recently, LLMs have been increasingly employed across a range of AM tasks \citep{li2025large, chen2024exploringpotentiallargelanguage}. LLMs have been used to assess argument quality and emotional appeal \citep{chen-eger-2025-emotions}, perform stance detection through reasoning chains \citep{ma2024chainstancestancedetection}, assist in dataset creation and summarization \citep{li-etal-2024-side,liu2023argugptevaluatingunderstandingidentifying}, and automate evaluation of argumentative outputs \citep{dhole-etal-2025-conqret}. These developments mark a shift from traditional supervised pipelines toward hybrid methods combining LLM reasoning with structured representations. Our work follows this direction by integrating LLM-based argument extraction into formal argumentation frameworks, linking natural language text to abstract argumentative graphs.

\section{LLM-Based Argument Mining}

Despite their impressive fluency, current LLMs still struggle to perform argument mining in an end-to-end, single-pass setting. This challenge is twofold: (i) LLMs often fail to maintain long-range structure and global consistency, and (ii) argument mining remains underspecified, as arguments can be represented in multiple ways. To address these challenges, we design a multi-stage pipeline that incrementally constructs an argumentative graph from raw text. Our framework incorporates not only basic components (premises and conclusions) and relations (support and attack), but also more nuanced phenomena studied in argumentation theory, such as implicit premises, linked premises, and undercuts \citep{walton1996argument, Freeman2011-FREASR-2, Kelley2014-KELTAO-10, walton2008argumentation, Walton_2008}.

\begin{figure}[t]
 \centering
 \includegraphics[width=1\textwidth]
{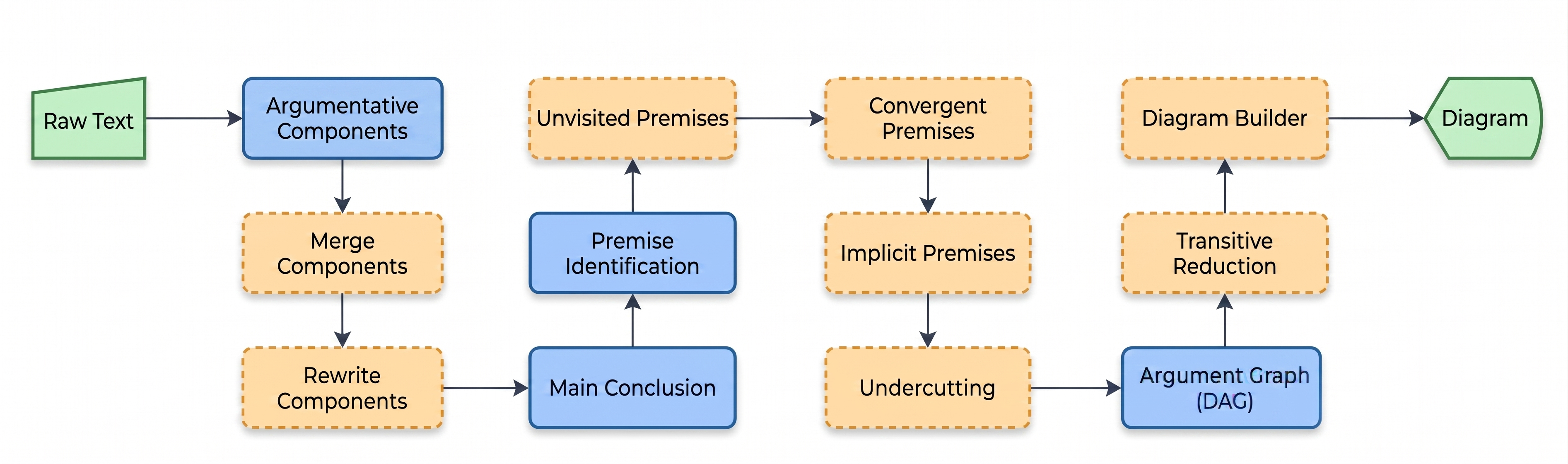}
 \caption{Overview of the system pipeline. The model converts natural language text into an argumentative directed acyclic graph. Blue boxes denote mandatory steps, while beige boxes denote optional steps.}
 \label{fig:pipeline}
\end{figure}

The pipeline combines LLM-driven extraction and analysis modules, coupled with an explicit representation schema. Each stage is implemented through targeted prompts to the LLM. Its modular design allows individual components to be independently disabled or replaced with alternative implementations. The following sections describe each stage of the pipeline, with optional steps marked by an \textit{asterisk} (*). Figure \ref{fig:pipeline} shows the full pipeline.

\paragraph{[1] Argumentative Components.}
The first step consists of identifying and enumerating the argumentative components within a text. The objective is to distinguish elements that contribute to the argument—such as claims and premises—from those that do not. The components identified at this stage constitute the foundational building blocks of the argument, with irrelevant content filtered out. Crucially, the role of a given textual span depends not only on its intrinsic content but also on its surrounding context \citep{opitz2019dissecting}. This context sensitivity provides LLMs with a distinct advantage: unlike rule-based and traditional machine learning approaches, which typically rely on surface-level features, LLMs can infer whether a span contributes to an argument by leveraging the rich contextual and discourse-level information acquired during pretraining.

\paragraph{[2] Merge Components.*} After identifying the individual argumentative components, the next step is to detect and merge those that are redundant or closely connected within the same line of reasoning. Merging involves analyzing the semantic and logical relationships among components to identify overlaps or dependencies. Components should be merged if they paraphrase one another, elaborate on the same point, or form a coherent reasoning chain—such as a conditional statement and its implication, or a component that is distributed across multiple sentences. This process ensures that the argumentation structure remains concise and logically unified, preventing fragmentation and enhancing interpretability. This step is particularly useful for longer arguments, but may be omitted for shorter ones or when a more fine-grained analysis is desired.

\paragraph{[3] Component Rewriting.*}
Once the individual components have been identified, they are rewritten for clarity. This process resolves incomplete or context-dependent expressions—such as pronouns or ellipses—by making each component self-contained and unambiguous.\footnote{In principle, this step could also be extended to summarize components that are overly long or complex, resulting in a more concise representation of the argument.} This step primarily improves readability and may be omitted when such refinement is not required.

\paragraph{[4] Conclusion Identification.} After extracting the argumentative components, the next step is to reconstruct the logical relations among them. As an initial step, the main conclusion is identified from the set of components.

\paragraph{[5] Premise Relation.} 
Given the main conclusion, support and attack relations between argumentative components are identified using a \textit{recursive} strategy. Specifically, the premises of each conclusion are determined, beginning with the main conclusion and progressively expanding to its supporting or attacking components. This decomposition reduces the complexity of the graph construction task, as only the premises for a single target conclusion need to be determined at a time, rather than inferring the entire structure at once. At each step, the system considers the full text, the list of argumentative components, and the current target conclusion, and returns the components that support or attack the target, or outputs \texttt{0} if no such components exist. Each component is expanded as a conclusion at most once. When a premise is identified for a given conclusion, it is added to a queue of components to be processed, following a breadth-first traversal of the argument graph. To ensure that the resulting structure is acyclic, candidate premises are restricted to unvisited components only. This prevents edges from pointing to ancestors or to components at the same hierarchical level in the partially constructed graph, thereby enforcing a directed acyclic graph structure.

\paragraph{[6] Check Unvisited Premises.*}
Because premise identification is performed locally for each target conclusion, some argumentative components may remain unassigned as premises and thus disconnected from the constructed argument graph. To address this, each unvisited component is examined to determine where it should be attached. Since unvisited components may also relate to one another, previously disconnected components can be assigned as mutual premises. When a cycle is detected, the components involved are merged into a single composite unit, and the attachment procedure is repeated for the resulting component. Although this step is optional, omitting it typically leads to missing components and a more fragmented graph.

\paragraph{[7] Linked and Convergent Premises.*} 
Following argumentation theory representations, relations between premises at the same hierarchical level are classified as either linked or convergent \citep{walton1996argument}. In a convergent relation, each premise provides an independent reason to support or attack the conclusion; in a linked relation, multiple premises jointly justify or refute the conclusion. Linked premises are represented through an intermediate empty node, capturing their joint contribution as a single inferential unit, rather than as independent supports. This step yields a more accurate representation of the inferential structure, albeit at the cost of an increased number of nodes in the resulting graph.

\paragraph{[8] Implicit Premises.*}
Arguments often omit premises that are necessary to fully support or attack a conclusion; such arguments are known as \textit{enthymemes} \citep{walton1987informal}. These premises may be left unstated for various reasons: they may be assumed to be shared by interlocutors, regarded as common knowledge, or omitted to draw attention to more contentious points. \textit{Implicit premises} are premises assumed in the argument but not explicitly expressed. Recovering them is essential for accurately reconstructing the structure of an argument, as they operate in tandem with explicit components to justify or refute a conclusion. As with explicit premises, we identify implicit ones recursively. For each inferential relation, we prompt the model to generate plausible implicit premises. In the case of convergent premises, this involves requesting common assumptions underlying the entire set. The newly generated components are added to the dictionary of argumentative elements, and the logical relations are updated accordingly to incorporate them. Although recovering implicit premises often clarifies inferential connections, it may also introduce trivial or self-evident statements, requiring subsequent filtering or interpretation.

\paragraph{[9] Rebuttal and Undercut.*}  In argumentation theory, two types of attacking relations are often distinguished: in which a premise (or set of premises) directly challenges a component, and \textit{undercuts}, in which a premise targets the inference to the conclusion itself \citep{pollock1987defeasible, pollock2001defeasible}.
Consider the argument: ``The lawn is wet. Therefore, it rained.'' A rebuttal would be: ``The lawn isn't wet at all,'' which directly challenges the truth of the premise. An undercutter would be: ``The sprinkler ran overnight,'' which accepts the premise but blocks the inference to the conclusion by offering an alternative explanation. To operationalize this distinction, we examine each attacking relation (e.g., 4 attacks 6') and present the model with all inference links involving the attacked conclusion (e.g., 2 supports 6' or 6 supports 1). If the model indicates that an attacking relation is an undercut, we insert an empty intermediate node into the original inference so that the attack targets this node rather than the conclusion directly. This step enables the explicit representation of challenges to inferential links, albeit at the cost of increased structural complexity in the resulting graph.

\paragraph{[10] Argumentative Graph.} 
An argument consists of propositions connected by logical relations. To represent this structure, we model the arguments as \textit{graphs}. Each node represents an argumentative component, and each edge denotes a directed relation between components, indicating that the source node either justifies or refutes the target node. The relation between a premise and a conclusion is unidirectional, and the graph must be acyclic. Allowing a premise to be supported by its own conclusion would violate the requirement of independent justification. Accordingly, we represent arguments as \textit{directed acyclic graphs} (DAGs), defined as $G = (N, E)$, where $N$ is the set of nodes and $E$ is the set of edges. 

\paragraph{[11] Transitive Reduction.*} 
A transitive reduction removes all edges that are not necessary to preserve reachability between nodes. In our context, this means eliminating direct connections between a premise and a conclusion when an indirect path already exists. 
The goal is to eliminate superfluous edges that clutter the graph without adding meaningful information. These redundant links are often a byproduct of the recursive construction: while the model may initially detect a connection between a premise and a conclusion, it may only later identify that the relation is mediated through intermediate steps. Transitive reduction thus helps clarify the underlying logical structure of the argument.


\paragraph{[12] Diagram.*} 
A common challenge in argument evaluation is the need for a global view of the structure. Some logical relations may appear plausible in isolation but prove inadequate when considered in a broader context. Relying solely on metric-based evaluations—such as component classification—can underestimate the importance of key argumentative connections. To address this, we provide a diagram builder that enables visual inspection of the argumentative graph. This facilitates manual refinement of the pipeline and more informed prompting. More importantly, a visual representation aligns with what people intuitively expect from argument analysis. For this reason, argument diagramming is widely used in argumentation theory textbooks \citep{reed2007argument}. Our diagram builder supports visual distinctions between explicit and implicit premises, supporting and attacking relations, convergent and independent premises, and rebuttals and undercuts.

\subsection{Example}
We illustrate our system with an argument about the prioritization of financial interests over academic merit in a college setting extracted from \citep{Sacrini2023}.

\begin{tcolorbox}[colback=gray!10, colframe=black!40, title=Input Text, sharp corners=south, boxrule=0.5pt]\label{argument_example}
The teacher will have to approve all students, since any failure would result in the students failing to drop out of the course, and the college's board does not want to miss out on receiving any tuition fees from these students. It could be different if concern for academic merit took precedence in this college, but clearly, it is not the case, as, at the departmental meeting, the director warned that the financial balance of the institution is the absolute priority for the semester.
\end{tcolorbox}

\begin{figure*}[t]
  \centering

  \begin{subfigure}[t]{0.48\textwidth}
    \vspace{0pt} 
    \footnotesize
    \raggedright
    \begin{enumerate}[leftmargin=*, label=\textbf{\arabic*.}, itemsep=3pt, topsep=0pt]
      \item The teacher must approve all students.
      \item Failing students would cause them to drop out of the course.
      \item The college board does not want to lose tuition revenue from these students.
      \item If academic merit took precedence, the situation would be different.
      \item Academic merit does not take precedence at this college.
      \item The director said at a departmental meeting that financial balance is the institution's absolute priority this semester.
      \item If the board does not want to lose tuition revenue from these students, it will take measures to prevent those students from failing (for example, by instructing or pressuring staff to pass them).
      \item The teacher is obliged or compelled to follow the board's directives/priorities, so will approve students when the board demands it.
    \end{enumerate}
  \end{subfigure}\hfill
  \begin{subfigure}[t]{0.48\textwidth}
    \vspace{0pt} 
    \centering
    \includegraphics[width=0.92\linewidth]{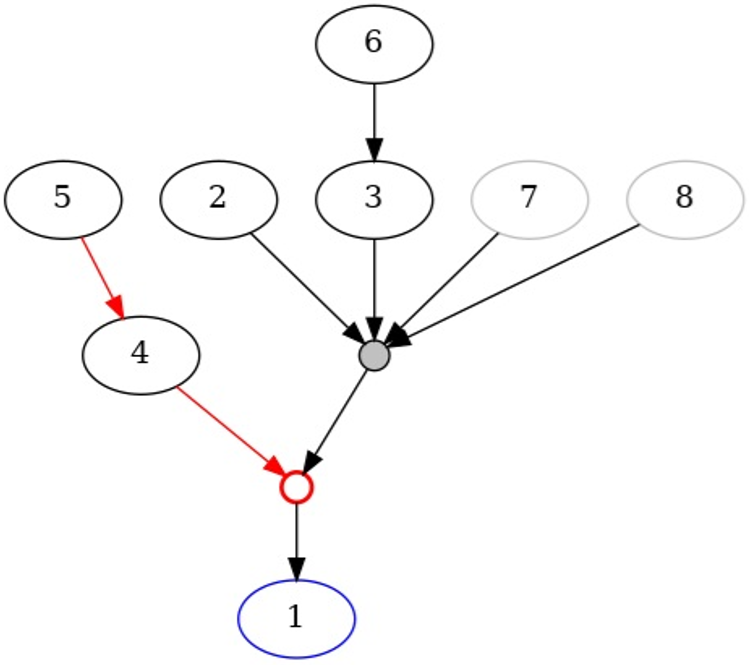}
  \end{subfigure}

  \newcommand{\argumentTeacher}{Teacher}

  \caption{Diagram of the \hyperref[argument_example]{\argumentTeacher} argument. Explicit premises are shown as \textit{black nodes}, and implicit premises as \textit{gray nodes}. \textit{Black edges} indicate support relations, and \textit{red edges} indicate attack relations. \textit{Small gray nodes} represent convergent premises, i.e., premises that jointly support or attack a conclusion. \textit{Small red nodes} represent undercutting attacks, targeting the inference rather than a specific premise. The \textit{blue node} indicates the conclusion.}
  \label{fig:teacher}
\end{figure*}

Figure \ref{fig:teacher} illustrates the reconstructed argument according to our full system. In the diagram, explicit premises are represented by \textit{black nodes}, while \textit{gray nodes} denote implicit premises. \textit{Black edges} indicate support relations, and \textit{red edges} indicate attack relations. \textit{Small gray nodes} represent convergent premises—i.e., premises that work together to support or attack a conclusion. \textit{Small red nodes} represent undercutting attacks, which target the inference itself rather than a specific premise. Finally, the \textit{blue node} indicates the conclusion of the argument.

\subsection{Data}
A key challenge in evaluating argument mining methods is the limited availability of annotated datasets. Compared to other NLP tasks, resources for argumentation remain relatively scarce \citep{peldszus2015annotated, reed2006preliminary, shnarch2020unsupervised}. Moreover, existing datasets differ substantially in their annotation schemes: they vary in the types of argumentative components they define (e.g., claim, premise, or more fine-grained roles such as proposal and observation) \citep{stab2017parsing, accuosto2020mining}, in the relations they distinguish (e.g., support, attack, evidential links) \citep{mayer2020transformer, park2018corpus}, and in the overall structures they assume (e.g., trees vs. graphs) \citep{rinott2015show, rocha2023assessing}. 


This lack of consistency makes comparing models across datasets challenging. In particular, none of these datasets includes the exact representations we employ, such as counterarguments. They also do not indicate implicit premises or convergent arguments. To address this issue, we constructed our own dataset from Introduction to Argumentative Analysis: Theory and Practice by Marcus Sacrini, an argumentation theory textbook \citep{Sacrini2023}. The dataset contains 42 short arguments on commonsense and philosophy topics, with an average length of 332.10 characters ($\sigma = 225.20$). The diagrams in the book use an annotation scheme similar to ours and were manually annotated by the author.\footnote{The diagrams also include additional features, such as logical-type support, which we did not incorporate.}


To ensure comparability, we selected two widely used argument mining datasets with representations relatively close to our graph-based model: AAEC \cite{stab2017parsing} and AbstRCT \cite{mayer2020transformer}. AAEC consists of persuasive essays written by students, whereas AbstRCT contains biomedical abstracts describing randomized controlled trials. Both datasets annotate argumentative components and relations in a way that is broadly compatible with our framework, enabling us to test our system with only minor adaptations. Both AAEC and AbstRCT include three component types—major claim, claim, and premise—as well as relations indicating support or attack. In addition, AbstRCT distinguishes between two types of attacking relations: attack and partial attack, the latter representing a weaker form of opposition. The datasets also differ in text length: arguments in AbstRCT are generally closer in size to the shorter texts in our dataset, whereas AAEC contains substantially longer documents.

For all experiments, we generated argument graphs for every instance in the dataset using three models: \texttt{GPT-4.1}, \texttt{GPT-5}, and \texttt{GPT-5-mini}. The latter offers a good compromise between output quality and cost. For each model, we generated a single set of graphs using a fixed random seed. Qualitative assessment revealed considerable variability in the generated argument structures across runs. Accordingly, generating multiple candidates and selecting the most coherent one may yield more robust results in practice.


\section{Experiments}
We conducted two experiments: (i) a manual evaluation on examples from an argumentation theory textbook (internal evaluation), and (ii) a quantitative evaluation on existing benchmark datasets (external evaluation).

\subsection{Internal Evaluation}
To assess the quality of the graphs produced by our method on the argumentation theory dataset, we adopt a three-level evaluation framework: component-, structure-, and global-level assessments, spanning from individual components to the full argumentative graph. The component level, which admits a more objective evaluation, is assessed through direct comparison between ground truth and predictions, while the structure and global levels rely on more open-ended, qualitative judgments. Detailed annotation criteria are provided in Appendix \ref{app:annotation}.

\subsubsection{Component Evaluation}
We first evaluated our system’s ability to identify basic argumentative elements—individual components, conclusions, and one-to-one relations. To this end, we compared automatically generated graphs with the ground truth diagrams. Because this evaluation was relatively straightforward, it was conducted by a single annotator. The following tasks were assessed quantitatively:

\begin{itemize}
\item \textbf{Span Detection.} The evaluation assessed whether the system correctly identified the text spans corresponding to argumentative components (claims and premises). Precision, recall, and F1-score were computed against the reference.
\item \textbf{Conclusion Detection.} Accuracy was calculated to determine how well the system identified the main conclusions, based on the annotated conclusions in the reference.
\item \textbf{Relation Detection.} Accuracy was used to assess whether directed links between components were correctly identified. Cases with fewer than two shared components between reference and generated graphs were excluded. The evaluation focused on premise–conclusion links, disregarding premise relations (e.g., convergence, independence), counterarguments, implicit premises, and support types.
\end{itemize}

The upper block of Table~\ref{tab:all_results} reports the outcomes of the comparison between automatically generated graphs and the gold-standard diagrams.  Minor wording differences—such as changes in tense, modal verbs, or voice—were intentionally ignored during evaluation, as they do not affect the argumentative content.

\paragraph{Discussion}
Results showed near-perfect accuracy (92.50\%) in conclusion detection. Occasional errors occurred when the system merged a premise and a conclusion or when the conclusion was implicit, something the system could not handle. For span detection, scores were lower (precision 80.31\%, recall 72.87\%, F1-score 75.70\%), likely due to annotation inconsistencies. The system often merged sentences that the gold standard had split into two, although both expressed the same meaning. In such cases, multiple valid annotations were possible, so these discrepancies did not necessarily indicate model error. For relation detection, the evaluation considered only arguments containing at least two matching components. Within this constrained setup, the generated and reference graphs aligned closely, achieving an accuracy of 80.57\%.

\begin{table*}[t!]
\centering
\caption{Evaluation results for the generated argument graphs. Component evaluations report precision, recall, F1-score, and accuracy, whereas structure and global evaluations report the mean score (1–3 scale) and exact agreement between annotators.}\label{tab:all_results}
\begin{tabular}{lcccccc}
\hline
\textbf{Task / Criterion} & \textbf{Precision} & \textbf{Recall} & \textbf{F1} & \textbf{Accuracy} & \textbf{Quality (\textit{EA})} \\
\hline
\multicolumn{6}{l}{\textbf{Component Evaluation}} \\
Span Detection     & 80.31 & 72.87 & 75.70 &  &     \\
Conclusion Detection  &  &  &   & 92.50 &   \\
Relation Detection  &  &  &   & 80.57 &   \\
\hline
\multicolumn{6}{l}{\textbf{Structure Evaluation}} \\
Implicit Premises             &   &   &  &   & 2.86 (0.95) \\
Undercuts              &   &   &  &   & 2.90 (0.88) \\
Convergent/Linked Premises &   &   &  &   & 2.85 (0.80) \\
\hline
\multicolumn{6}{l}{\textbf{Global Evaluation}} \\
Completeness &   &   &  &   & 2.96 (0.97) \\
Faithfulness &   &   &  &   & 2.92 (0.88) \\
\hline
\end{tabular}
\end{table*}

\subsubsection{Structure Evaluation}
A second set of evaluations focused on the structural patterns of the arguments—that is, whether the components were correctly organized. In particular, the analysis examined the appropriate use of implicit premises, undercuts, and the structural organization of premises (convergent or linked). Annotators had access to the original text, and reference diagrams were used as a point of comparison, although alternative interpretations were accepted. To assess these structural aspects, two independent annotators were employed. The evaluation criteria were as follows:

\begin{itemize}
\item \textbf{Implicit Premises.} For each inferential relation, annotators assessed whether implicit premises were adequate, missing, or redundant.
\item \textbf{Undercuts.} Annotators evaluated whether undercuts were correctly identified.
\item \textbf{Convergent and Linked Premises.} Annotators judged whether the premises were appropriately classified as linked (joint support) or convergent (independent support).
\end{itemize}

Appendix~\ref{app:annotation} provides a detailed description of the annotation criteria for this and the subsequent evaluation.

\paragraph{Discussion}
The middle block of Table~\ref{tab:all_results} presents the results of these evaluations. Each criterion was rated on a 1–3 scale, and both the mean score across annotators and their rate of exact agreement (EA) were reported.\footnote{Cohen’s kappa was not reported, as it tends to be unreliable when score distributions are highly skewed.} Overall, the arguments were rated as very good across all criteria, with a high level of agreement among annotators: Implicit Premises (M = 2.86, EA = 0.95), Undercuts (M = 2.90, EA = 0.88), and Convergent/Linked Premises (M = 2.85, EA = 0.80). The evaluation of undercuts, however, was less reliable, as only eight arguments included them. In these cases, much of the reasoning relied on counterfactual judgments rather than explicit textual evidence.

\subsubsection{Global Evaluation}
Finally, a set of global evaluations was conducted to capture overall properties of the generated argument graphs. These criteria provided a holistic assessment of whether the graphs adequately covered the main argumentative content while remaining faithful to the original text. As in the previous evaluation, two annotators independently performed the assessments.

\begin{itemize}
\item \textbf{Completeness.} Assessed whether the graph captured all major claims and premises expressed in the text.
\item \textbf{Faithfulness.} Assessed whether the graph avoided introducing information not supported by the text (i.e., no hallucinated nodes or edges).
\end{itemize}

\paragraph{Discussion}
The bottom block of Table~\ref{tab:all_results} summarized the evaluations of completeness and faithfulness. Both criteria were rated on a 1–3 scale, and the mean score and exact agreement between annotators were reported. Overall, the results were very good for both Completeness (M = 2.96, EA = 0.97) and Faithfulness (M = 2.92, EA = 0.88). These scores indicated that the generated graphs generally captured the main argumentative structures of the texts while maintaining high fidelity to their content. The strong agreement between annotators further suggests that the overall quality of the generated argument graphs was consistent and reliable.

\subsection{External Evaluation}

To analyze the quality of our LLM-based framework on public datasets, we employ the simplest representation, as to permit comparison with those different formats. Thus, we turn off the steps indicated by an asterisk in our pipeline --- i.e., steps 1, 4, 5, 6.\footnote{We do not enforce the DAG constraint and therefore disable step 10 either, as doing so would require merging cyclic components, which in turn would entail modifying the original components.} Also, as mentioned above, both AAEC and AbstRCT classify components into three types: major claim, claim, and premise, whereas we only use conclusion and premise. Thus, we map components to these categories based on the structure of the generated graphs: the root node is considered the major claim, components directly attached to it are treated as claims, and all remaining lower-level elements are considered premises. As for the partial-attack relation from the AbstRCT dataset, we developed a special prompt to detect this as a third type of relation.

We evaluate the generated argument structures using standard argument mining tasks commonly adopted in the literature \citep{morio2022end}. Although our framework produces complete argument graphs, breaking the evaluation down into these tasks enables direct comparison with existing systems. Task definitions and evaluation protocols are detailed below:

\begin{itemize} 
\item \textbf{Span Identification.}
This task consists of detecting portions of the input text corresponding to argumentative components. A component is considered correctly identified if it exactly matches a ground-truth component. Performance is evaluated using the (micro) \textit{F1-score} over the predicted and ground-truth spans.

\item \textbf{Component Classification.}
This task assigns a component type to each span according to the datasets' annotation scheme (premise, claim, or major claim). Performance is evaluated using \textit{F1} and \textit{Macro-F1} scores over the component labels.

\item \textbf{Relation Classification.}
This task determines the argumentative relations between pairs of components and involves two aspects. First, the model predicts whether a relation exists between two components, regardless of its type. This evaluates the argument graph structure and is reported as \textit{Link} using \textit{F1}. Second, the model assigns a label to each detected relation (e.g., support or attack). Relation type classification is evaluated using \textit{F1} and \textit{Macro-F1} scores.
\end{itemize}




\begin{table*}[t!]
\centering
\caption{Results on the AbstRCT and AAEC datasets. The full pipeline includes the argument component identification step, whereas the gold setting uses ground-truth components.}
\label{tab:datasets}
\begin{tabular}{lcccccc}
\hline
\textbf{Dataset / Model} 
& \multicolumn{1}{c}{\textbf{Span}} 
& \multicolumn{2}{c}{\textbf{Component}} 
& \multicolumn{3}{c}{\textbf{Relation}} \\[-0.1em]
\cmidrule(lr){2-2} \cmidrule(lr){3-4} \cmidrule(lr){5-7} \\[-1.2em]
& \textbf{F1} 
& \textbf{F1} & \textbf{Macro} 
& \textbf{Link} & \textbf{F1} & \textbf{Macro} \\
\hline

\multicolumn{7}{l}{\textbf{AbstRCT}} \\
LLM-based pipeline (full)
& 59.65 & 27.64 & 19.26 & 20.82 & 19.33 & 8.8 \\

\citet{morio2022end}
& \textbf{70.93} & \textbf{64.78} & \textbf{44.56}
& 39.74 & 38.71 & \textbf{33.94} \\

LLM-based pipeline (gold comp.)
& - & 61.93 & 38.53 & \textbf{51.51} & \textbf{46.21} & 23.45 \\

\hline
\multicolumn{7}{l}{\textbf{AAEC}} \\

LLM-based pipeline (full)
& 33.03 & 22.70 & 13.26 & 0.41 & 0.34 & 0.19 \\

\citet{morio2022end}
& \textbf{85.20} & \textbf{75.66} & \textbf{67.03}
& \textbf{55.72} & \textbf{55.17} & \textbf{41.92} \\

LLM-based pipeline (gold comp.)
& - & 62.64 & 54.15 & 22.88 & 22.67 & 23.11 \\

\hline
\end{tabular}
\end{table*}

Table \ref{tab:datasets} reports the results of this experiment. The full pipeline (first row) yields poor overall performance. This is largely because the model is not designed to select argument spans under strict criteria, resulting in low overlap with the ground truth annotations. Since the evaluation requires exact span matching, even minor deviations lead to substantial performance drops in all tasks, as errors propagate (see Appendix~\ref{sec:threshold_analysis} for a detailed analysis).

The third row shows the results obtained when gold-standard components are provided to the system. Performance improves substantially, indicating that the main bottleneck lies in span detection rather than in component or relation classification. Under these conditions, our approach achieves its best results on AbstRCT, reaching $51.51$ in link prediction and $46.21$ in relation detection—improvements of $+11.77$ and $+7.5$ over the state-of-the-art baseline. This suggests that the model effectively captures relational structure once the components are known. Performance is also stronger on AbstRCT than on AAEC, likely due to its shorter texts, similar in length to the examples from the argumentation theory textbook.

\section{Conclusion}
Argumentation frameworks and LLMs exhibit complementary limitations in argument mining: methods based on argumentation frameworks often struggle with the complexity and variability of real-world arguments, whereas LLM-based approaches lack systematic and theoretically grounded standards for argument reconstruction. In this paper, we proposed an LLM-based system for reconstructing argumentative structures from natural language text. The system leverages the commonsense knowledge of LLMs to recover arguments while imposing a standardized graph-based representation. Our results show that the system can adequately reconstruct argumentation-theory-style arguments and, when adapted to different annotation schemes, achieves reasonable performance across benchmark datasets. At the same time, the results highlight persistent limitations in component identification and the need for improved prompting strategies. Future work will focus on evaluating the system on more complex and realistic datasets, refining the representation to capture richer argumentative phenomena (e.g., premise types and argument schemes), and incorporating evaluative criteria derived from argumentation theory.

\section*{Acknowledgment} 
The authors of this work would like to thank the Center for Artificial Intelligence (C4AI-USP) and the support from the São Paulo Research Foundation (FAPESP grant \#2019/07665-4) and from the IBM Corporation. F. G. C. was partially supported by CNPq grants 312180/2018-7 and 305753/2022-3. The authors also thank support by CAPES -- Finance Code 001. The authors are grateful to Marcus Sacrini for allowing the use of examples from his book in the construction of the dataset.


\bibliography{iclr2026_conference}
\bibliographystyle{iclr2026_conference}

\clearpage

\appendix

\section{Diagrams}\label{app:diagrams}
We present additional examples of argument graphs from our dataset. Nodes represent argumentative components: black nodes indicate explicit premises, while gray nodes denote implicit premises. Edges encode relations, with black edges representing support and red edges indicating attacks. Smaller gray nodes correspond to convergent premises. Undercutting relations—i.e., attacks on inferences—are shown as red edges terminating in red nodes. The conclusion is highlighted in blue.

\begin{figure*}[h]
  \centering
  \includegraphics[width=\textwidth]{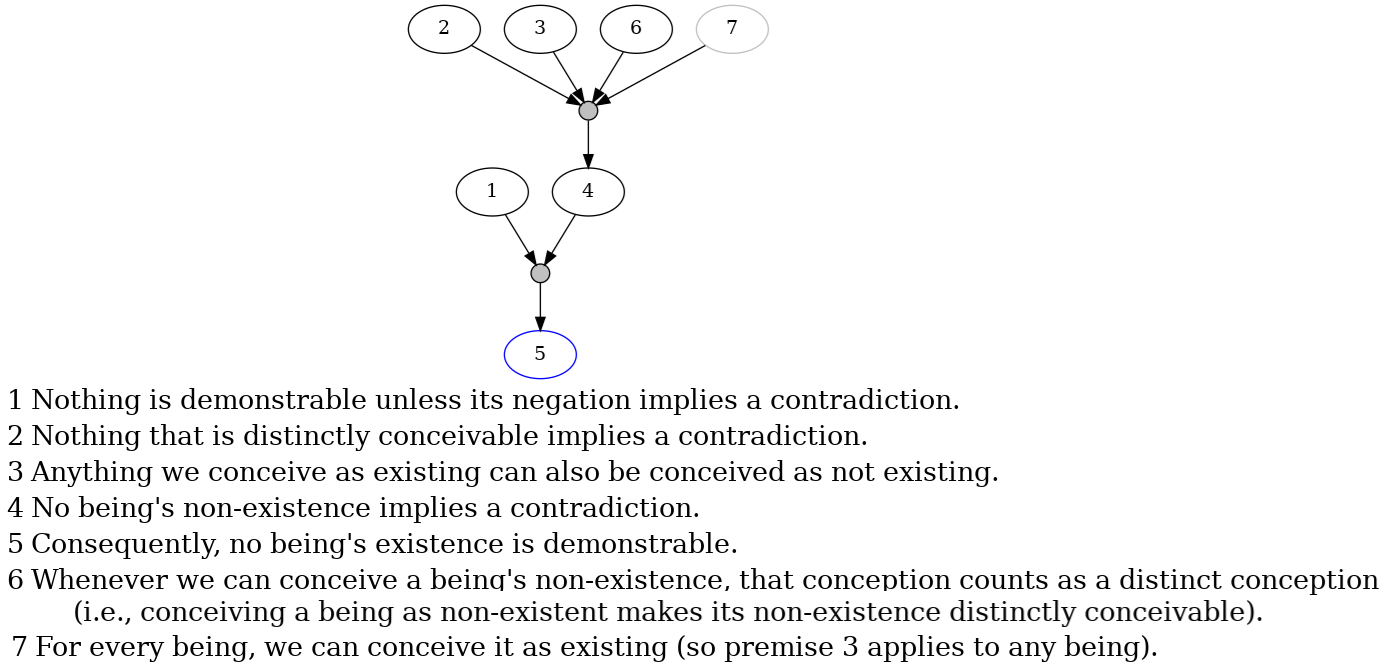}
  \caption{Text: ``Nothing is demonstrable, unless the contrary implies a contradiction. Nothing, that is distinctly conceivable, implies a contradiction. Whatever we conceive as existent, we can also conceive as non-existent. There is no being, therefore, whose non-existence implies a contradiction. Consequently there is no being, whose existence is demonstrable. I propose this argument as entirely decisive, and am willing to rest the whole controversy upon it.''}
  \label{fig:hume_being}
\end{figure*}

\begin{figure*}[h]
  \centering
  \includegraphics[width=\textwidth]{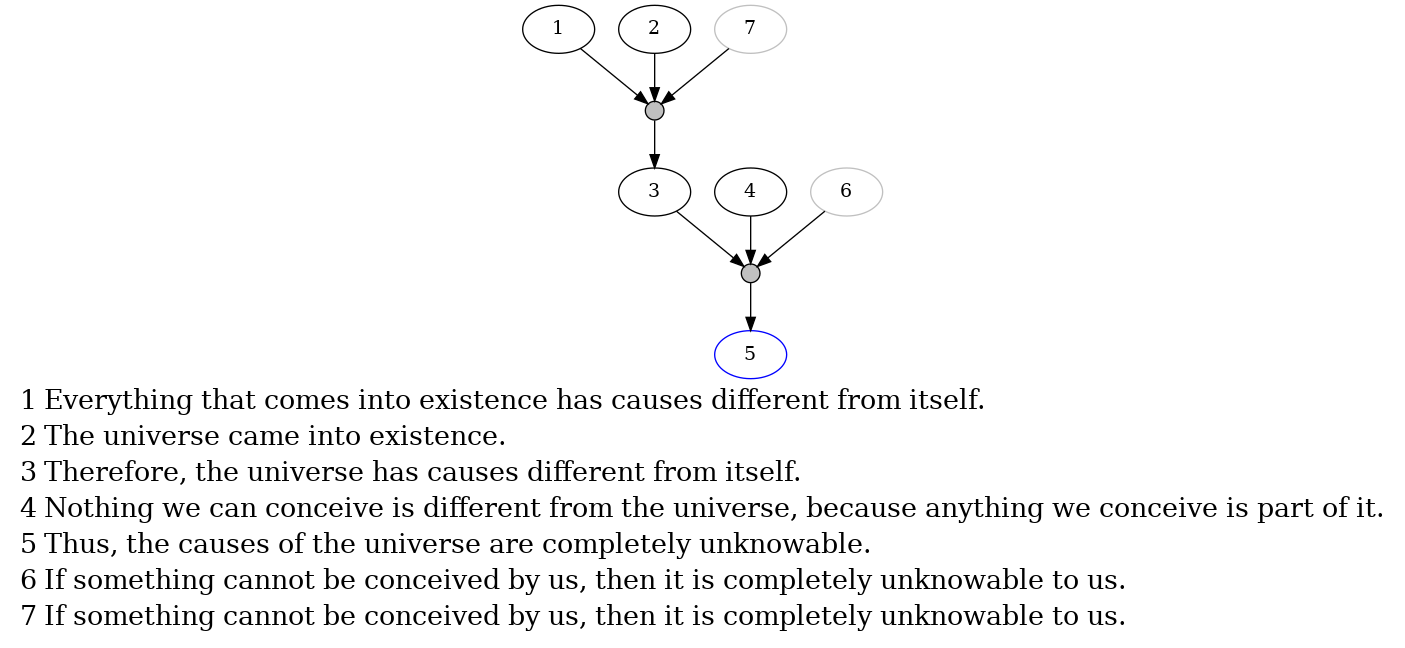}
  \caption{Text: ``Everything that comes into existence has causes different from itself. The universe came into existence. Therefore, it has causes different from itself. However, nothing we can conceive of is different from the universe, for everything we can conceive of as something is already part of the universe. Thus, the causes of the universe are completely unknowable.''}
  \label{fig:universe}
\end{figure*}

\begin{figure*}[h]
  \centering
  \includegraphics[width=\textwidth]{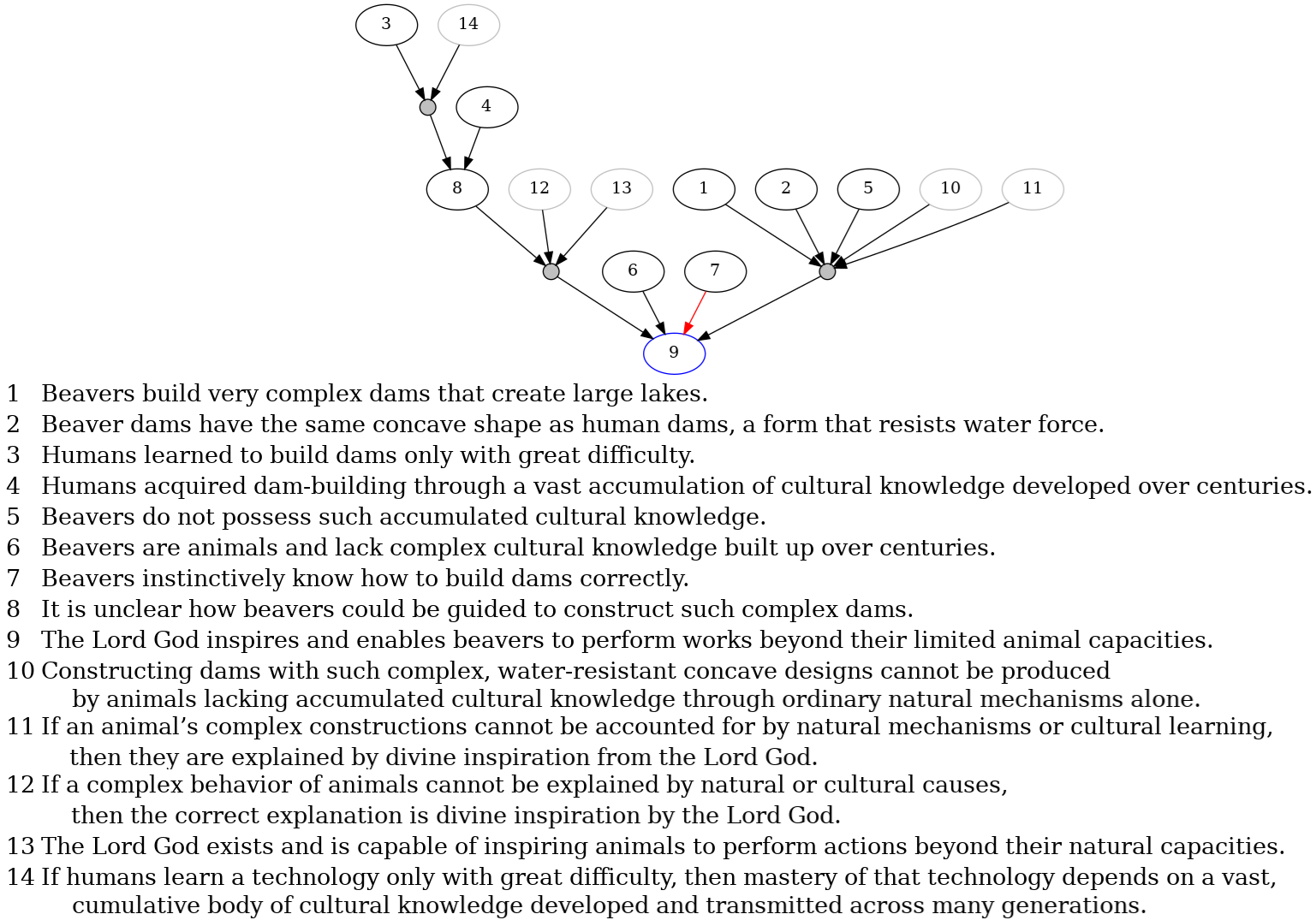}
  \caption{Text: ``Beavers build very complex dams that create large lakes. These dams are built in the same concave shape as those constructed by humans — one of the shapes that best resists the force of water. However, humans learned to build dams only with great difficulty. They owe this ability to a vast accumulation of cultural knowledge developed and consolidated over centuries. This is not the case with beavers. They are animals that do not possess complex cultural knowledge built up over centuries. Beavers simply know how to build dams correctly. How, then, could they be guided to perform constructions of such complexity? Clearly, it is the Lord God who inspires them and allows them to carry out works so far beyond their limited animal capacities.''}
  \label{fig:universe}
\end{figure*}

\begin{figure*}[h]
  \centering
  \includegraphics[width=\textwidth]{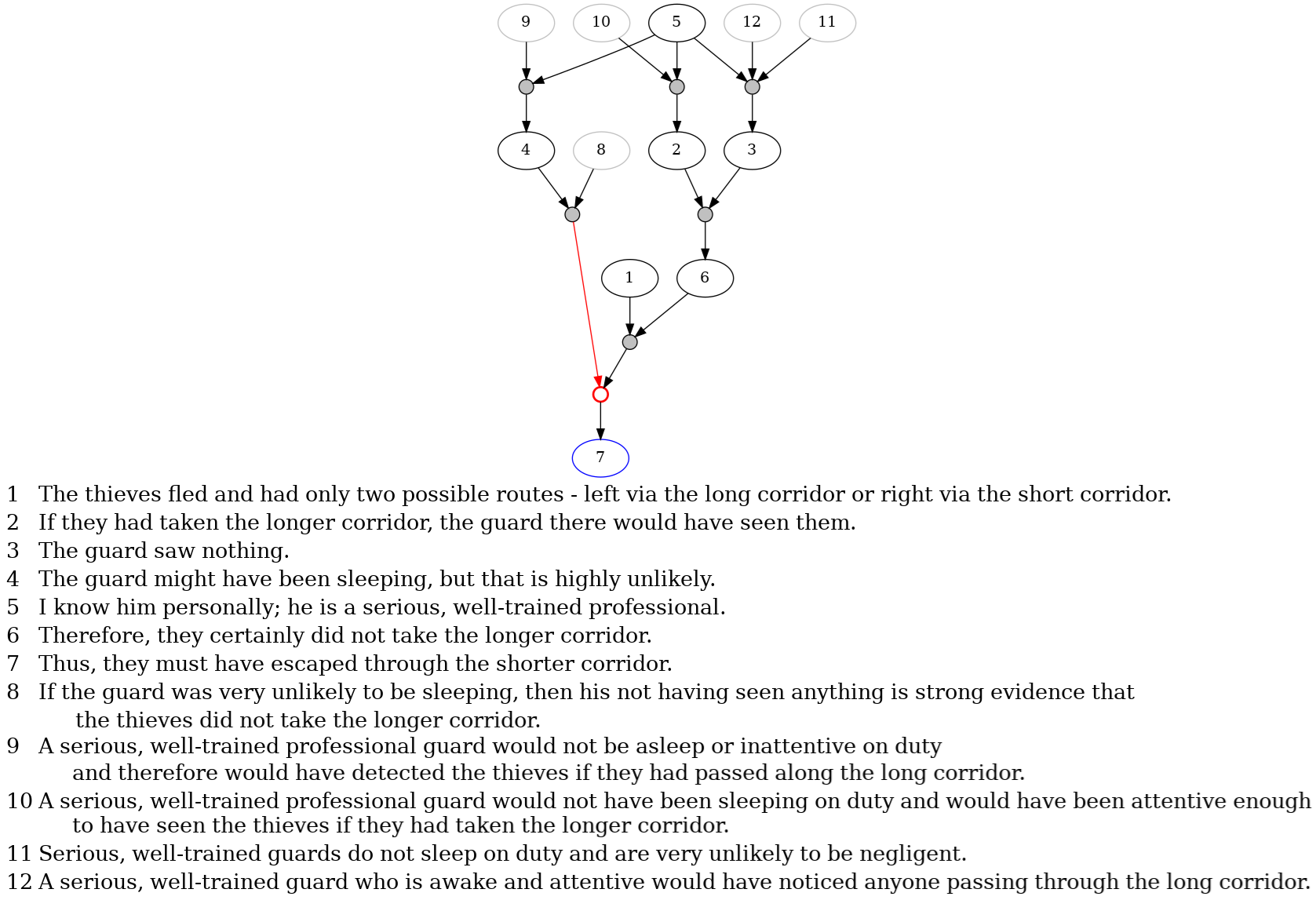}
  \caption{Text: ``The thieves fled and there are only two paths they could have taken — to the left, through the long corridor, or to the right, through the short corridor. If they had taken the longer corridor, they would have been seen by the guard who was there. But the guard saw nothing. It might be possible that the guard was sleeping, but that is highly unlikely. I know him personally, I know he is a serious and well-trained professional. Therefore, it is certain they did not take the longer corridor. Thus, they must have escaped through the shorter corridor.''}
  \label{fig:thieves}
\end{figure*}

\clearpage

\section{Annotation Criteria}\label{app:annotation}
In this section, we describe how we organized the methodology for annotation across the different evaluation stages. 

For the first evaluation, referred to as Component Evaluation, we enlisted a colleague with experience in argumentation mining. The annotator assessed whether each pair of components—ground truth and prediction—referred to the same content, disregarding minor differences in wording.

For the subsequent evaluations—Structure and Global—we employed two independent evaluators. We began by presenting the evaluation criteria and explaining in detail how each task should be performed. Then, we provided five representative argumentative examples, which had been personally annotated by the authors, so that the evaluators could apply the criteria in practice. After completing this initial annotation exercise, the evaluators returned their annotations and shared questions and feedback, allowing us to identify ambiguities and refine the guidelines. Once the methodology was finalized, the evaluators proceeded to annotate the complete set of arguments following the revised instructions.

Evaluators followed a set of well-defined criteria, each assessed on a three-point scale (1--3). They had access to the original text, as well as to the reference diagrams. They were instructed to use the latter as a point of comparison, with the understanding that alternative interpretations were possible. The criteria were designed to capture both the internal quality of the argument structure and its faithfulness to the original text. Table~\ref{tab:criteria} summarizes these criteria and their corresponding scales.

\begin{table}[H]
\centering
\small
\setlength{\tabcolsep}{3pt}
\renewcommand{\arraystretch}{1.25}
\begin{tabularx}{\columnwidth}{p{0.45\columnwidth} X}
\hline
\textbf{Criterion} & \textbf{Scale} \\
\hline

\textbf{Implicit Premises}\\
Evaluates whether the argument includes all necessary premises and whether they are well justified.
&
\parbox[t]{\hsize}{
1 = crucial premises missing\\
2 = redundant or vague\\
3 = complete and well justified
} \\

\hline
\textbf{Counter-Arguments}\\
Assesses the correctness of counter-arguments in the argument.
&
\parbox[t]{\hsize}{
1 = incorrect or absent\\
2 = partially correct (e.g., should be a direct attack rather than a counter-argument)\\
3 = correct and complete (or not required)
} \\

\hline
\textbf{Convergent/Divergent Premises}\\
Verifies whether convergent and divergent premises are correctly identified according to the text.
&
\parbox[t]{\hsize}{
1 = incorrect\\
2 = partially correct\\
3 = correct in relation to the text
} \\

\hline
\textbf{Completeness}\\
Measures the extent to which the argument includes all central claims and premises.
&
\parbox[t]{\hsize}{
1 = important parts missing\\
2 = partially complete\\
3 = covers all main claims and core premises
} \\

\hline
\textbf{Fidelity}\\
Evaluates the accuracy of the argument’s content in relation to the source text.
&
\parbox[t]{\hsize}{
1 = contains information not present in or distorts the original text\\
2 = generally faithful but with minor additions or modifications\\
3 = fully faithful to the text
} \\
\hline
\end{tabularx}
\caption{Criteria used for the Structure and Global Evaluations.}
\label{tab:criteria}
\end{table}

\clearpage

\section{Analysis of Component Similarity and Its Impact on Performance}\label{sec:threshold_analysis}

\begin{figure}[h]
\centering
\includegraphics[width=0.8\linewidth]{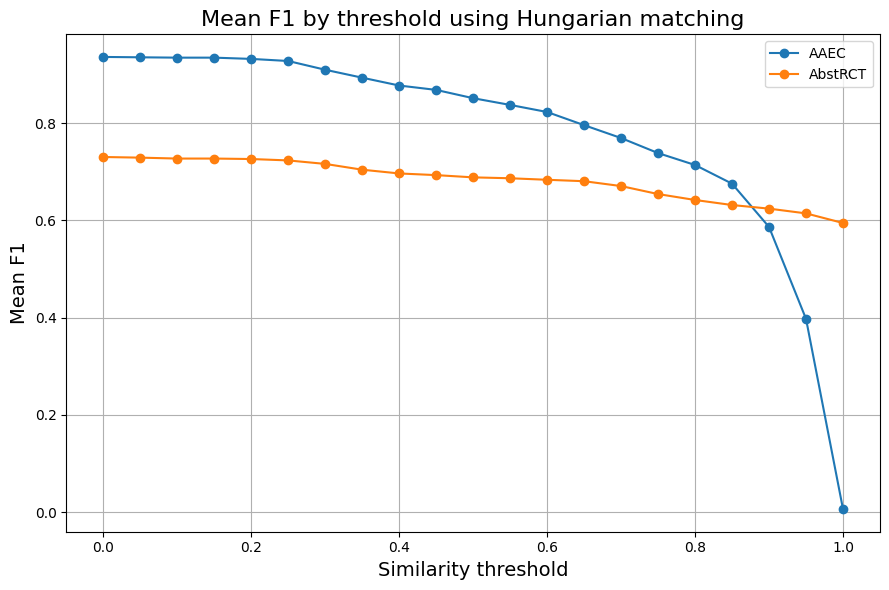}
\caption{Mean F1-score as a function of the similarity threshold between predicted and gold components, using Hungarian matching.}
\label{fig:threshold_analysis}
\end{figure}

To investigate the performance degradation observed in Table~\ref{tab:datasets}, we analyze similarity between predicted and gold components independently of exact span matching. Figure~\ref{fig:threshold_analysis} reports the mean F1-score as a function of a similarity threshold, using optimal (Hungarian) matching and character-level overlap.

At moderate thresholds (e.g., $0.6$), performance remains relatively high (around $0.82$ for AAEC and $0.68$ for AbstRCT), indicating substantial lexical overlap between predictions and gold annotations. Even at a stricter threshold ($0.8$), F1-scores remain above $0.60$, suggesting that predicted components are often closely aligned in surface form.

However, performance drops sharply at higher thresholds. For AAEC, the F1-score declines rapidly beyond $0.9$, approaching zero at $1.0$, indicating that exact matches are rare despite high similarity. A similar, though less pronounced, trend is observed for AbstRCT. This highlights a limitation of strict evaluation protocols, where small boundary deviations invalidate otherwise strong matches.

These results support the hypothesis that component identification is the main bottleneck. The model frequently produces spans that are highly similar to the gold annotations but fail to meet exact matching criteria, likely due to the specificity of annotation guidelines and the absence of explicit constraints in the prompting strategy.

This analysis also explains the gap between the full pipeline and the gold-component setting in Table~\ref{tab:datasets}: when exact spans are provided, relation prediction improves significantly. Thus, the issue lies not in understanding argumentative content, but in aligning generated spans with annotation standards.

Overall, improving component identification—via better prompting, post-processing, or alignment—offers the greatest potential for enhancing end-to-end performance.

\end{document}